\pdfoutput=1

\documentclass[11pt]{article}

\usepackage{acl}

\usepackage{times}
\usepackage{latexsym}

\usepackage{amssymb}
\usepackage{amsmath}
\usepackage{enumitem}
\setlist{nosep}
\usepackage{tabularx}
\usepackage{multicol}
\usepackage{multirow}

\usepackage{booktabs}
\usepackage{comment}
\usepackage{caption}
\usepackage{subcaption}
\usepackage{enumitem}

\usepackage{tikz}
\usepackage{tikz-qtree}

\newcommand{\smallsep}{0.75em}
\newcommand{\midsep}{1.5em}

\usetikzlibrary{arrows,matrix,graphs,shapes,decorations,automata,backgrounds,petri,positioning,calc,decorations.pathmorphing,external,shapes.multipart}
\newcommand{\fixh}[1]{\raisebox{0pt}[0pt][0pt]{#1}}

\usepackage[T1]{fontenc}

\usepackage{microtype}
\title{Geometry-Aware Supertagging\\ with Heterogeneous Dynamic Convolutions}

\author{Konstantinos Kogkalidis \and Michael Moortgat\\
		Utrecht Institute of Language Sciences \\
  	    \texttt{k.kogkalidis,m.j.moortgat@uu.nl}
}

\begin{document}
\maketitle

\begin{abstract}
The syntactic categories of categorial grammar formalisms are structured units made of smaller, indivisible primitives, bound together by the underlying grammar's category formation rules.
In the trending approach of constructive supertagging, neural models are increasingly made aware of the internal category structure, which in turn enables them to more reliably predict rare and out-of-vocabulary categories, with significant implications for grammars previously deemed too complex to find practical use.
In this work, we revisit constructive supertagging from a graph-theoretic perspective, and propose a framework based on heterogeneous dynamic graph convolutions, aimed at exploiting the distinctive structure of a supertagger's output space.
We test our approach on a number of categorial grammar datasets spanning different languages and grammar formalisms, achieving substantial improvements over previous state of the art scores.
\end{abstract}

\section{Introduction}
Their close affinity to logic and the lambda calculus has made categorial grammars a standard tool of the trade for the formally-inclined NLP practitioner.
Modern flavors of categorial grammar, despite their (sometimes striking) divergences, share a common architecture.
At its core, a categorial grammar is a formal system consisting of two parts.
First, there is a \textit{lexicon}, a mapping that assigns to each word a set of \textit{categories}. 
Categories are quasi-logical formulas recursively built out of atomic categories by means of category forming operations.
The inventory of category forming operations at the minimum has the ability to express linguistic function-argument structure. 
If so desired, the inventory can be extended with extra operations, e.g.~to handle syntactic phenomena beyond simple concatenation, or to express additional layers of grammatical information.
The second component of the grammar is a small set of \textit{inference rules}, formulated in terms of the category forming operations. 
The inference rules dictate how categories interact and, through this interaction, how words combine to form larger phrases. 
Parsing thus becomes a process of deduction comparable (or equatable, depending on the grammar's formal rigor) to program synthesis, providing a clean and elegant syntax-semantics interface.

In the post-neural era, these two components allow differentiable implementations.
The fixed lexicon is replaced by \textit{supertagging}, a process that contextually decides on the most appropriate supertags (i.e. categories), whereas the choice of which rules of inference to apply is usually deferred to a parser further down the processing pipeline.
The highly lexicalized nature of categorial grammars thus shifts the bulk of the weight of a parse to the supertagging component, as its assignments and their internal make-up inform and guide the parser's decisions.

In this work, we revisit supertagging from a geometric angle.
We first note that the supertagger's output space consists of a sequence of trees, which has as of yet found no explicit representational treatment.
Capitalizing on this insight, we employ a framework based on heterogeneous dynamic graph convolutions, and show that such an approach can yield substantial improvements in predictive accuracy across categories both frequently and rarely encountered during a supertagger's training phase.

\section{Background}
\label{sec:background}
The supertagging problem revolves around the design and training of a function tasked with mapping a sequence of words $\{w_1, \dots, w_n\}$ to a sequence of categories $\{c_1, \dots, c_n\}$.
Existing supertagging architectures differ in how they implement this mapping, with each implementation choice boiling down to (i) which of the temporal and structural dependencies within and between the input and output are taken into consideration, and (ii) how these dependencies are materialized.

Earlier work would utilize solely occurrence counts from a training corpus to independently map word n-grams to their most likely categories, and then attempt to filter out implausible sequences via rule-constrained probabilistic models~\cite{bangalore-joshi-1999-supertagging}. 
The shift from sparse feature vectors to distributed word representations facilitated integration with neural networks and improved generalization on the mapping domain, extending it to rare and previously unseen words~\cite{lewis-steedman-2014-improved}.
Later, the advent of recurrent neural networks offered a natural means of incorporating temporal structure, widening the input receptive field through contextualized word representations on the one hand~\cite{xu-etal-2015-ccg}, but also permitting an auto-regressive formulation of the output generation, whereby the effect of a category assignment could percolate through the remainder of the output sequence~\cite{vaswani-etal-2016-supertagging}.
Regardless of implementation specifics, the discriminative paradigm employed by all above works fails to account for the smallness of the data; exceedingly rare categories are practically impossible to learn, and categories absent from the training data are completely ignored.

As an alternative, the recently emerging constructive paradigm seeks to explore the structure hidden \textit{within} categories.
By inspecting their formation rules,~\citet{kogkalidis-etal-2019-constructive} equates categories to CFG derivations, and views a category sequence as the concatenation of their flattened depth-first projections.
The goal sequence is now incrementally generated on a symbol-by-symbol basis using a transformer-based seq2seq model; a twist which provides the decoder with the means to construct novel categories on demand, bolstering co-domain generalization.
The decoder's global receptive field, however, comes at the heavy price of quadratic memory complexity, which also bodes poorly with the elongated output sequences, leading to a slowed down inference speed.
Expanding on the idea,~\citet{prange-etal-2021-supertagging} explicates the categories' tree structure, embedding symbols based on their tree positions and propagating contextualized representations through tree edges, using either residual dense connections or a tree-structured GRU.
This adaptation completely eliminates the burden of learning \textit{how} trees are constructed, instead allowing the model to focus on \textit{what} trees to construct, leading to drastically improved performance.
Simultaneously, since the decoder is now token-separable, it permits construction of categories for the entire sentence in parallel, speeding up inference and reducing the network's memory footprint.
In the process, however, it loses the ability to model interactions between auto-regressed nodes belonging to different trees, morally reducing the task once more to sequence classification (albeit now with a dynamic classifier).

Despite their common goal of accounting for syntactic categories in the zipfian tail, there are tension points between the above two approaches.
In providing a global history context, the first breaks the input-to-output alignment and hides the categorial tree structure.
In opting for a tree-wise bottom-up decoding, the second forgets about meaningful \textit{inter}-tree output-to-output dependencies.
In this paper, we seek to resolve these tension points with a novel, unified and grammar-agnostic supertagging framework based on heterogeneous dynamic graph convolutions.
Our architecture combines the merits of explicit tree structures, strong auto-regressive properties, near-constant decoding time, and a memory complexity that scales with the input, boasting high performance across the full span of the frequency spectrum and surpassing previously established benchmarks on all datasets considered.

\section{Methodology}

\subsection{Breadth-First Parallel Decoding}
Despite seeming at odds, both architectures described fall victim to the same trap of conflating problem-specific structural biases and general purpose decoding orders: one forgets about tree structure in opting for a sequential decoding, whereas the other does the exact opposite, forgetting about sequential structure in opting for a tree-like decoding.
We note first that the target output is (a batch of) neither sequences nor trees, but rather \textit{sequences of trees}.
Having done that, our task is of a purely technical nature: we simply need to come up with the spatiotemporal dependencies that abide by \textit{both} structural axes, and then a neural architecture that can accommodate them.

\citet{prange-etal-2021-supertagging} make a compelling case for depth-parallel decoding, given that it's incredibly fast (i.e., not temporally bottlenecked by left-to-right sequential dependencies) but also structurally elegant (trees are only built when/if licensed by non-terminal nodfes, ensuring structural correctness virtually for free). 
Sticking with depth-parallel decoding means necessarily foregoing some autoregressive interactions: we certainly cannot look to the future (i.e., tree nodes located deeper than the current level, since these should depend on the decision we are about to make), but neither to the present (i.e., tree nodes residing in the current level, since these will be all decided simultaneously).
This still leaves some leeway as to what could constitute the prediction context.
The maximalist position we adopt here is nothing less than the entire past, i.e. \textit{all} the nodes we have so far decoded.
Crucially, this extends beyond the ancestry-bound ``vertical interactions'' of a tree unfolding function implemented \textit{{\`a} la} treeRNN, allowing ``diagonal'' interactions between autoregressed nodes living in different trees.

Such exotic interactions do not follow the inductive biases of any run-of-the-mill architecture, forcing us to turn our attention to structure-aware dynamic convolutions.
To make the architecture conducive to learning while keeping its memory footprint in check, we repurpose the encoder's word vectors from initial seeds to recurrent state-tracking vectors that arbitrate the decoding process across both \textit{sequence length} and \textit{tree depth}, respecting the ``regularly irregular'' structure of the output space.
In high level terms, the process can be summarized as an iteration of three alternating stages of message passing rounds.
\begin{enumerate}
	\item State vectors are initialized by some external encoder.
	\item An empty fringe consisting of blank nodes is instantiated, one such node per word, rooting the corresponding supertag trees.
	\item Until a fix-point is reached (there is no longer any fringe):
		\begin{enumerate}
			\item States project class weights to their respective fringe nodes in a one-to-many fashion. Depending on the arity of the decoded symbols, a next fringe of unfilled nodes is constructed at the appropriate positions.
			\item Each state vector receives feedback in a many-to-one fashion from the just decoded nodes above (what used to be the fringe), yielding tree-contextual states.
			\item The updated state vectors emit and receive messages to one another in a many-to-many fashion, yielding tree- and sequence- contextual states.
		\end{enumerate}
\end{enumerate}
For a visual rendition, refer to Appendix~\ref{app:viz}.

% Initially, state vectors are supplied from an arbitrary encoder network, and a fringe consisting of $s$ unlabeled nodes is instantiated in alignment with the input sequence.
% From then on, and until a fix-point is reached:
% \begin{enumerate}
%     \item Each state vector receives feedback in a many-to-one fashion from the last decoded nodes lying directly above it (initially none), yielding tree-contextual states.
%     \item  The updated state vectors exchange messages with one another in a many-to-many fashion,
%     yielding tree-and-sequence-contextual states.
%     \item The final states project class weights to their respective fringe nodes in a one-to-many fashion; depending on the arity of the decoded symbols, a next masked fringe is constructed with appropriate node positions and state-to-node edge indices; the process terminates when the next fringe is empty.
% \end{enumerate}

\subsection{Architecture}
We now move on to detail the individual blocks that together make up the network's pipeline.

\subsubsection{Node Embeddings}
State vectors are temporally dynamic; they are initially supplied by an external encoder, and are then updated through a repeated sequence of three message passing rounds, described in the next subsections.
Tree nodes, on the other hand, are not subject to temporal updates, but instead become dynamically ``revealed'' by the decoding process. 
Their representations are computed on the basis of (i) their primitive symbol and (ii) their position within a tree.

Primitive symbol embeddings are obtained from a standard embedding table $W_e: \mathcal{S} \to \mathbb{R}^{d_n}$ that contains a distinct vector for each symbol in the set of primitives $\mathcal{S}$. 
When it comes to embedding positions, we are presented with a number of options.
It would be straightforward to fix a vocabulary of positions, and learn a distinct vector for each.
Such an approach would however lack elegance, as it would impose an ad-hoc bound to the shape of trees that can be encoded (contradicting the constructive paradigm), while also failing to account for the compositional nature of trees.
We thus opt for a path-based approach, inspired by and improving upon the idea of~\citet{shiv2019novel}.
We note first that \textit{paths} over binary branching trees form a semi-group, i.e. they consist of two primitives (namely a left and a right path), and an associative non-commutative binary operator that binds two paths together into a single new one.
The archetypical example of a semigroup is matrix multiplication; we therefore instantiate a tensor $P \in \mathbb{R}^{2 \times n_d \times n_d}$ encoding each of the two path primitives as a linear map over symbol embeddings.
From the above we can derive a function $p$ that converts positions to linear maps, by performing consecutive matrix multiplications of the primitive weights, as indexed by the binary word of a node's position; e.g. the linear map corresponding to position $12_{10} = 1100_{2}$ would be $p(12) = P_0P_0P_1P_1 \in \mathbb{R}^{d_n \times d_n}$.
We flatten the final map by evaluating it against an initial seed vector $\rho_0$, corresponding to the tree root.%
\footnote{In practice, paths are efficiently computed once per batch for each unique tree position during training, and stored as fixed embeddings during inference.}
To stabilize training and avoid vanishing or exploding weights, we model paths as \textit{unitary} transformations by parameterizing the two matrices of $P$ to orthogonality using the exponentiation trick on skew-symmetric bases~\cite{bader2019computing,lezcano2019trivializations}.

The representation $n_{i, k}$ of a tree node $\sigma \in \mathcal{S}$ occupying position $k$ in tree $i$ will then be given as the element-wise product of its tree-positional and content embeddings:
\[
n_{i, k} = p(k)(\rho_0) \odot \left(W_e(\sigma)\right) \in \mathbb{R}^{d_n}
\]
The embedder is then essentially an instantiation of a binary branching unitary RNN~\cite{arjovsky2016unitary}, where the choice of which hidden-to-hidden map to follow at each step depends on the node's position relative to its ancestor.%
	\footnote{Concurrently, \citet{bernardy2022assessing} follow a similar approach in teaching a unitary RNN to recognize Dyck words, and find the unitary representations learned to respect the compositional properties of the task.
	Here we go the other way around, using the unitary recurrence exactly because we expect them to respect the compositional properties of the task.}
Since paths are shared across trees, their representations are in practice efficiently computed once per batch for each unique tree position during training, and stored as fixed embeddings during inference.

\subsubsection{Node Prediction}
Assuming at step $\tau$ a sequence of globally contextualized states $h^{\tau}$, we need to use each element $h^{\tau}_i$ to obtain class weights for all of the node neighborhood $\mathcal{N}_{i, \tau}$ consisting of all nodes (if any) of tree $i$ that lie at depth $\tau$.
We start by down-projecting the state vector into the node's dimensionality using a linear map $W_n$.
The resulting feature vectors are indistinguishable between all nodes of the same tree -- to tell them apart (and obtain a unique prediction for each), we gate the feature vectors against each node's positional embedding.
From the latter, we obtain class weights by matrix multiplying them against the transpose of the symbol embedding table~\cite{press-wolf-2017-using}:
\[
\mathrm{weights}_{i,k} = \left(p(k)(\rho_0) \odot W_n h^{\tau}_i\right) W_e^\top
\]
The above weights are converted into a probability distribution over the alphabet symbols $\mathcal{S}$ by application of the $\mathrm{softmax}$ function. 

\subsubsection{Autoregressive Feedback}
\label{subsec:nf}
We update states with information from the last decoded nodes using a heterogeneous message-passing scheme based on graph attention networks~\cite{velivckovic2018graph,brody2021attentive}.
First, we use a bottleneck layer $W_b$ to down-project the state vector into the nodes' dimensionality.
For each position $i$ and corresponding state $h^\tau_i$, we compute a self-loop score:
\[
\tilde{\alpha}_{i,\circlearrowleft,\tau} = w_{a} \cdot (W_b(h^\tau_i) \ ||  \ \mathbf{0})
\]
where $w_a \in \mathbb{R}^{2d_n}$ a dot-product weight and $\mathbf{0}$ a $d_n$-dimensional zero vector.
Then we use the (now decoded) neighborhood $\mathcal{N}_{i, \tau}$ to generate a heterogeneous attention score for each node $n_{i, k} \in \mathcal{N}_{i, \tau}$:
\[
\tilde{\alpha}_{i,k,\tau} = w_a \cdot (h^{\tau}_i \ || \ n_{i, k})
\]
Scores are passed through a leaky rectifier non-linearity before being normalized to attention coefficients $\alpha$.
These are used as weighting factors that scale the self-loop and input messages, the latter upscaled by a linear map $W_m$:
\[
	\tilde{h}^\tau_i = \sum_{n_{i, k} \in \mathcal{N}_{i,\tau}} {\alpha}_{i,k,\tau}W_m n_{i,k} + {\alpha}_{i,\circlearrowleft,\tau} h^\tau_i
\]
This can also be seen as a dynamic residual connection -- $\alpha_{i,\circlearrowleft,\tau}$ acts as a gate that decides how open the state's representation should be to node feedback (or conversely, how strongly it should retain its current values).
States receiving no node feedback (i.e. states that have completed decoding one or more time steps ago) are thus protected from updates, preserving their content.
In practice, attention coefficients and message vectors are computed for multiple attention heads independently, but these are omitted from the above equations to avoid cluttering the notation.

\subsubsection{Sequential Feedback}
At the end of the node feedback stage, we are left with a sequence of locally contextualized states $\tilde{h}^\tau_i$.
The sequential structure can be seen as a fully connected directed graph, nodes being states (words) and edges tabulated as the square matrix $\mathcal{E}$, with entry $\mathcal{E}_{i, j}$ containing the relative distance between words $i$ and $j$.
%$\mathcal{E}$ labeled by relative distances between word pairs.
We embed these distances into the encoder's vector space using an embedding table $W_r \in \mathbb{R}^{2\kappa \times d_w}$, where $\kappa$ the maximum allowed distance, a hyper-parameter.
Edges escaping the maximum distance threshold are truncated rather than clipped, in order to preserve memory and facilitate training, leading to a natural segmentation of the sentence into (overlapping) chunks.
Following standard practices, we project states into query, key and value vectors, and compute the attention scores between words $i$ and $j$ using relative-position weighted attention~\cite{shaw-etal-2018-self}:
\[
\tilde{a}_{i,j} = d_w^{-1/2}~(W_q \tilde{h}^\tau_i \odot W_r\mathcal{E}_{i,j}) \cdot W_k\tilde{h}^\tau_j
\]
From the normalized attention scores we obtain a new set of aggregated messages:
\[
m_{i,t}' = \sum_{j \in \{0..s\}} \frac{\mathrm{exp}(\tilde{a}_{i,j}) W_v\tilde{h}^\tau_j}{
\sum_{k\in \{0..s\}}
\mathrm{exp}(\tilde{a}_{i,k})}
\]
Same as before, queries, keys, values, edge embeddings and attention coefficients are distributed over many heads. 
Aggregated messages are passed through a swish-gated feed-forward layer~\cite{dauphin2017language,shazeer2020glu} to yield the next sequence of state vectors:
\[
h^{\tau+1}_i = W_3\left(\mathrm{swish}_1(W_1 m_{i,\tau}')\odot W_2m_{i,\tau}'\right)
\]
where $W_{1,2}$ are linear maps from the encoder's dimensionality to an intermediate dimensionality, and vice versa for $W_3$.

\subsubsection{Putting Things Together}
\label{subsubsec:ptt}
We compose the previously detailed components into a single layer, which acts a sequence-wide, recurrent-in-depth decoder.
We insert skip connections between the input and output of the message-passing and feed-forward layers~\cite{he2016deep}, and subsequently normalize each using root mean square normalization~\cite{zhang2019root}.

\section{Experiments}
\label{sec:experiments}
We employ our supertagging architecture in a range of diverse categorial grammar datasets spanning different languages and underlying grammar formalisms.
In all our experiments, we bind our model to a monolingual BERT-style language model used as an external encoder, fine-tuned during training~\cite{devlin2018bert}.
In order to homogenize the tokenization between the one directed by each dataset and the one required by the encoder, we make use of a simple localized attention aggregation scheme.
The subword tokens together comprising a single word are independently projected to scalar values through a shallow feed-forward layer.
Scalar values are softmaxed within their local group to yield attention coefficients over their respective BERT vectors, which are then summed together, in a process reminiscent of a cluster-wide attentive pooling~\cite{li2016gated}.
In cases of data-level tokenization treating multiple words as a single unit (i.e. assigning one type to what BERT perceives as many words), we mark all words following the first with a special \texttt{[MWU]} token, signifying they need to be merged to the left.
This effectively adds an extra output symbol to the decoder, which is now forced to do double duty as a sequence chunker.
To avoid sequence misalignments and metric shifts during evaluation, we follow the merges dictated by the ground truth labels, and consider the decoder's output as correct only if all participating predictions match, assuming no implicit chunking oracles.

\subsection{Datasets}
We conduct experiments on the two variants of the English CCGBank, the French TLGbank and the Dutch \AE thel proofbank.
A high-level overview of the datasets is presented in Table~\ref{tab:datasets}, and short descriptions are provided in the following paragraphs.
We refer the reader to the corresponding literature for a more detailed exposition.

\begin{table}[h]
    \centering
    \small{
    \begin{tabularx}{1\linewidth}{@{}l@{~}c@{\quad}c@{\quad}c@{\quad}c@{}}
        & \multicolumn{2}{c}{\textbf{\textit{CCGbank}}} 
        & \textbf{\textit{TLGbank}} 
        & \textbf{\textit{\AE thel}}\\ 
        & \textit{original} & \textit{rebank} \\
        \toprule
        \textbf{Primitives}     & 37        & 40        & 27        & 81\\
        ~ Zeroary           & 35        & 38        & 19        & 31\\ 
        ~ Binary            & 2         & 2         & 8         & 50\\
        \midrule
        \textbf{Categories}     & 1323      & 1619      & 851      & 5762\\
        ~{in train}         & 1286      & 1575      & 803      & 5146\\
        ~{depth avg.}       & 1.94      & 1.96      & 1.99     & 1.82   \\
        ~{depth max.}       & 6         & 6         & 7        & 35\\ 
        \midrule
        \textbf{Test Sentences} & 2407      & 2407      & 1571     & 5770\\
        ~{length avg.}      & 23.00     & 24.27     & 27.58    & 16.52 \\
        \midrule
        \textbf{Test Tokens}    & 55371     & 56395     & 44302     & 95331\\
        ~ Frequent {(100+)}   & 54825     & 55690     & 43289     & 91503\\
        ~ Uncommon {(10-99)}  & 442       & 563       & 833       & 2639\\
        ~ Rare {(1-9)}        & 75        & 107       & 149       & 826\\
        ~ Unseen {(OOV)}      & 22        & 27        & 31        & 363\\
    \end{tabularx}}
    \caption{Bird's eye view of datasets employed and relevant statistics. Test tokens are binned according to their corresponding categories' occurrence count in the respective dataset's training set. Token counts are measured before pre-processing. Unique primitives for the type-logical datasets are counted after binarization.}
    \label{tab:datasets}
\end{table}

\paragraph{CCGBank}
The English CCGbank (original)~\cite{hockenmaier2007ccgbank} and its refined version (rebank)~\cite{honnibal2010rebanking} are resources of Combinatory Categorial Grammar (CCG) derivations obtained from the Penn Treebank~\cite{taylor2003penn}.
CCG~\cite{steedman2011combinatory}
builds lexical categories with the aid of two binary slash operators, capturing forward and backward function application. Some additional rules lent from combinatory logic~\cite{curry1958combinatory} permit constrained forms of type raising and function composition, allowing categories to remain relatively short and uncomplicated while keeping parsing complexity in check.
\begin{comment}
The English CCGbank (original)~\cite{hockenmaier2007ccgbank} and its refined version (rebank)~\cite{honnibal2010rebanking} are resources of Combinatory Categorial Grammar (CCG) proofs derived from the Penn Treebank~\cite{taylor2003penn}.
CCG~\cite{steedman2011combinatory}
builds lexical categories with the aid of two binary operators, meant to capture a directed notion of function application, and employs derivational rules lent from combinatory logic logic~\cite{curry1958combinatory} that permit constrained forms of type raising and function composition outside of the lexical level, allowing its categories to remain relatively short and uncomplicated while keeping its parsing complexity in check.
\end{comment}
The key difference between the two versions lies in their tokenization and the plurality of categories assigned, the latter containing more assignments and a more fine-grained set of syntactic primitives, which in turn make it a slightly more challenging evaluation benchmark.

\paragraph{French TLGbank}
The French type-logical treebank~\cite{moot2015type} is a collection of proofs extracted from the French treebank~\cite{abeille2003building}.
The theory underlying the resource is that of Multi-Modal Typelogical Grammars~\cite{moortgat96multimodal}; annotations are deliberately made compatible with Displacement Calculus~\cite{morrill2011displacement} and First-Order Linear Logic~\cite{moot2001linguistic} at the cost of a small increase in lexical sparsity.
In short, the vocabulary of operators is extended with two modalities that find use in licensing or restricting the applicability of rules related to non-local syntactic phenomena.
To adapt their representation to our framework, we cast unary operators into pseudo-binaries by inserting an artificial terminal tree in a fixed slot within them.
Due to the absence of predetermined train/dev/test splits, we randomize them with a fixed seed at a 80/10/10 ratio and keep them constant between repetitions.

\paragraph{\AE thel}
Our last experimental test bed is \AE thel~\cite{kogkalidis-etal-2020-aethel}, a dataset of type-logical  proofs for written Dutch sentences, automatically extracted from the Lassy-Small corpus~\cite{noord2013large}.
\AE thel is geared towards {\em semantic} parsing, which means categories employ linear implication $\multimap$ as their single binary operator. 
An additional layer of dependency information is realized via unary modalities, now lifted to \textit{classes} of operators distinguishing complement and adjunct roles.
The grammar assigns concrete instances of polymorphic coordinator types, as a result containing more and sparser categories (some of which distinctively tall); considering also its larger vocabulary of primitives, it makes for a good stress test for our approach.
We experiment with the latest available version of the dataset (version \texttt{1.0.0a5} at the time of writing).
Same as before, we impose a regular tree structure, this time by merging adjunct (resp. complement) markers with the subsequent (resp. preceding) binary operator, which makes for an unambiguous and invertible representational translation.

\begin{table*}[h]
    \newcolumntype{N}{>{\centering\arraybackslash}p{0.089\textwidth}}
    \newcolumntype{L}{>{\raggedright\arraybackslash}X}
    \newcommand{\num}[1]{\multirow{1}{*}{#1}}
    \newcommand{\ave}[2]{\multirow{1}{*}{~~#1}{\textsubscript{$\pm$#2}}}

    \centering
    {\small
    \begin{tabularx}{1.00\textwidth}{@{}l@{~}XN@{}N@{}N@{}N@{}N@{}}
    & & \multicolumn{5}{c}{\textbf{accuracy} {(\%)}} \\
    \cmidrule(lr){3-7}
    & \multicolumn{1}{c}{\textbf{model}} & {overall} & {frequent} & {uncommon} & {rare} & {unseen}\\
    \toprule
    \multicolumn{7}{l}{\textit{\textbf{CCG (original)}}} \\
    & \multicolumn{1}{l}{Symbol Sequential LSTM /w n-gram oracles
    {~\cite{liu2021generating}}}
    & 95.99 & 96.40 & 65.83 & 
    \multicolumn{2}{c}{8.65\textsuperscript{!}} \\
    & \multicolumn{1}{l}{Cross-View Training
    {~\cite{clark-etal-2018-semi}}} 
    & 96.10 & -- & -- & -- & n/a \\ 
    & \multicolumn{1}{l}{{Recursive Tree Addressing}
    {~\cite{prange-etal-2021-supertagging}}}
    &  \num{96.09} & \num{96.44} & \num{68.10} & \num{\textbf{37.40}} & \num{\textbf{3.03}} \\
    & \multicolumn{1}{l}{{BERT Token Classification}
    {~\cite{prange-etal-2021-supertagging}}}
    & \num{96.22} & \num{\textbf{96.58}} & \num{70.29} & \num{23.17} & n/a\\
    & \multicolumn{1}{l}{Attentive Convolutions
    {~\cite{tian-etal-2020-supertagging}}} 
    & \textbf{96.25} & \textbf{96.64} & 71.04 & n/a & n/a\\ 
    \addlinespace
    & \multicolumn{1}{l}{{Heterogeneous Dynamic Convolutions}
    {~(this work)}} 
    & \ave{\textbf{96.29}}{0.04}
    & \ave{\textbf{96.61}}{0.04}
    & \ave{\textbf{72.06}}{0.72}
    & \ave{34.45}{1.58}
    & \ave{\textbf{4.55}}{2.87}\\
    \addlinespace
    \addlinespace
    %%%%%%%%%%%%%%%%%%%%%%%%%%%%%%%%%%%%%%%%%%%%%%%%%%%%%%%%%%%%%%%%%%%%%%%%%%%%%%%%%%%%%%%%%%%%%%%%%%%%%%%%%
    \multicolumn{2}{L}{\textit{\textbf{CCG (rebank)}}} \\
    & \multicolumn{1}{l}{{Symbol Sequential Transformer}\textsuperscript{\textdagger}
    {~\cite{kogkalidis-etal-2019-constructive}}}
    & \num{90.68} & \num{91.10} & \num{63.65} & \num{34.58} & \num{\textbf{7.41}} \\
    & \multicolumn{1}{l}{{TreeGRU}
    {~\cite{prange-etal-2021-supertagging}}}
    & \num{94.62} & \num{95.10} & \num{64.24} & \num{25.55} & \num{2.47} \\
    & \multicolumn{1}{l}{{Recursive Tree Addressing}
    {~\cite{prange-etal-2021-supertagging}}}
    &  \num{94.70} & \num{95.11} & \num{68.86} & \num{\textbf{36.76}} & \num{4.94} \\
    & \multicolumn{1}{l}{{Token Classification}
    {~\cite{prange-etal-2021-supertagging}}}
    &  \num{94.83} & \num{95.27} & \num{68.68} & \num{23.99} & \num{n/a} \\
    \addlinespace
    & \multicolumn{1}{l}{{Heterogeneous Dynamic Convolutions}
    {~(this work)}} 
    & \ave{\textbf{95.07}}{0.04}
    & \ave{\textbf{95.45}}{0.04}
    & \ave{\textbf{71.40}}{1.15}
    & \ave{\textbf{37.19}}{1.81} 
    & \ave{{3.70}}{0.00}\\
    \addlinespace
    \addlinespace
    %%%%%%%%%%%%%%%%%%%%%%%%%%%%%%%%%%%%%%%%%%%%%%%%%%%%%%%%%%%%%%%%%%%%%%%%%%%%%%%%%%%%%%%%%%%%%%%%%%%%%%%%%
    \multicolumn{7}{l}{\textit{\textbf{French TLGbank}}} \\
    % \midrule
    & \multicolumn{1}{l}{{ELMo \& LSTM Classification}
    {~\cite{moot2019}}}
    & \num{93.20} & \num{95.10} & \num{75.19} & \num{25.85} & \num{n/a}\\
    & \multicolumn{1}{l}{{BERT Token Classification\textsuperscript{\textdaggerdbl}}}
    & \num{\textbf{95.93}} & \num{\textbf{96.44}} & \num{81.39} & \num{47.45} & \num{n/a}\\
    \addlinespace
    & \multicolumn{1}{l}{{Heterogeneous Dynamic Convolutions}
    {~(this work)}}
    & \ave{\textbf{95.92}}{0.01}
    & \ave{96.40}{0.01} 
    & \ave{\textbf{81.48}}{0.97}
    & \ave{\textbf{55.37}}{1.00} 
    & \ave{\textbf{7.26}}{2.67} \\ 
    \addlinespace
    \addlinespace
    %%%%%%%%%%%%%%%%%%%%%%%%%%%%%%%%%%%%%%%%%%%%%%%%%%%%%%%%%%%%%%%%%%%%%%%%%%%%%%%%%%%%%%%%%%%%%%%%%%%%%%%%%   
    \multicolumn{7}{l}{\textit{\textbf{\AE thel}}} \\
    % \midrule
    & \multicolumn{1}{l}{{Symbol Sequential Transformer}\textsuperscript{\textborn}
    {~\cite{kogkalidis-etal-2020-neural}}}
    & \num{83.67} & \num{84.55} & \num{64.70} & \num{50.58} & \num{\textbf{24.55}} \\
    & \multicolumn{1}{l}{{BERT Token Classification}\textsuperscript{\textdaggerdbl}}
    &  \num{93.52} & \num{\textbf{94.83}} & \num{71.85} & \num{38.06} & \num{n/a} \\
    \addlinespace
    & \multicolumn{1}{l}{{Heterogeneous Dynamic Convolutions}
    {~(this work)}}
    & \ave{\textbf{94.08}}{0.02}
    & \ave{\textbf{95.16}}{0.01}
    & \ave{\textbf{75.55}}{0.02}
    & \ave{\textbf{58.15}}{0.01}
    & \ave{18.37}{2.73} \\
    \bottomrule
    \addlinespace
    \multicolumn{7}{l}{{\fixh{\textsuperscript{!}}%
    Accuracy over both bins, with a frequency-truncated training set (authors claim no difference when using the full set).}}\\
    \multicolumn{7}{l}{{\textsuperscript{\textdagger}Numbers from~\citet{prange-etal-2021-supertagging}.}}\\
    \multicolumn{7}{l}{{\fixh{\textsuperscript{\textdaggerdbl}}Our replication.}}\\
    \multicolumn{7}{l}{{\textsuperscript{\textborn}Model trained and evaluated on an older dataset version and tree sequences spanning less than 140 nodes in total.}}
    \end{tabularx}}
    \caption{Model performance across datasets and compared to recent studies. Numbers are taken from the papers cited unless otherwise noted. For our model, we report averages and standard deviations over 6 runs. Bold face fonts indicate (within standard deviation of) highest performance.}
    \label{tab:comparisons}
\end{table*}

\subsection{Implementation}
We implement our model using PyTorch Geometric~\cite{Fey/Lenssen/2019}, which provides a high-level interface to efficient low-level protocols, facilitating fast and pad-free graph manipulations.
We share a single hyper-parameter setup across all experiments, obtained after a minimal logarithmic search over sensible initial values.
Specifically, we set the node dimensionality $d_n$ to 128 with 4 heterogeneous attention heads and the state dimensionality $d_w$ to 768 with 8 homogeneous attention heads.
We train using AdamW~\cite{loshchilov2018fixing} with a batch size of 16, weight decay of $10^{-2}$, and a learning rate of $10^{-4}$, scaled by a linear warmup and cosine decay schedule over 25 epochs.
During training we provide strict teacher forcing and apply feature and edge dropout at 20\% chance.
Our loss signal is derived as the label-smoothed negative log-likelihood between the network's prediction and the ground truth label~\cite{muller2019does}.
We procure pretrained base-sized BERT variants from the transformers library~\cite{wolf-etal-2020-transformers}: RoBERTa for English~\cite{liu2019roberta}, BERTje for Dutch~\cite{de2019bertje} and CamemBERT for French~\cite{martin-etal-2020-camembert}, which we fine-tune during training, scaling their learning rate by 10\% compared to the decoder.

\subsection{Results}
We perform model selection on the basis of validation accuracy, and gather the corresponding test scores according to the frequency bins of Table~\ref{tab:datasets}.
Table~\ref{tab:comparisons} presents our results compared to relevant published literature.
Evidently, our model surpasses established benchmarks in terms of overall accuracy, matching or surpassing the performance of both traditional supertaggers on common categories and constructive ones on the tail end of the frequency distribution.

We observe that the relative gains appear to scale with respect to the task's complexity.
In the original version of the CCGbank, our model is only slightly superior to the next best performing model (in turn only marginally superior to the token-based classification baseline), whereas in the rebank version the absolute difference is one order of magnitude wider.
The effect is even further pronounced for the harder type-logical datasets, which are characterized by a longer tail,
leading to performance comparable to CCGbank's for the French TLGbank (despite it being significantly smaller and sparser), and a 10\% absolute performance leap for \AE thel (despite its unusually tall and complex types).
We attribute this to increased returns from performance in the rare and uncommon bins; there is a synergistic effect between the larger population of these bins pronouncing even minor improvements, and acquisition of rarer categories apparently benefiting from the plurality of their respective bins in a self-regularizing manner.
Put simply, learning sparse categories is \textit{easier} and \textit{matters more} for grammars containing many rare categories.

Finally, to investigate the relative impact of each network component, we conduct an ablation study where message passing components are removed from their network in their entirety.
Removing the state feedback component collapses the network into a token-wise separable recurrence, akin to a graph-featured RNN without a hidden-to-hidden affine map.
Removing the node feedback component turns the network into a Universal Transformer~\cite{dehghani2018universal} composed with a dynamically adaptive classification head.
Removing both is equatable to a 1-to-many contextualized token classification that is structurally unfolded in depth.
Our results, presented in Table~\ref{tab:ablations}, verify first a positive contribution from both components, indicating the importance of both information sharing axes.
In three out of the four datasets, the relative gains of incorporating state feedback outweigh those of node feedback, and are most pronounced in the case of \AE thel, likely due to its positionally agnostic types.
With the exception of CCGrebank, relinquishing both kinds of feedback largely underperforms having either one, experimentally affirming their compatibility.

\begin{table}
    \centering
    {\small
    \begin{tabularx}{0.8\linewidth}{@{}l@{\qquad}ccc@{}}
        &  {-sf}  & {-nf} & { -sf-nf} \\
        \toprule
        \textit{\textbf{CCG} (original)} & -0.05 & -0.01 & -0.08 \\ 
        \textit{\textbf{CCG} (rebank)} & -0.12 & -0.04 & -0.07 \\
        \textit{\textbf{French TLGbank}} & -0.13 & -0.14 & -0.23 \\
        \textit{\textbf{\AE thel}} & -0.24 & -0.12 & -0.37
    \end{tabularx}}
    \caption{Absolute difference in overall accuracy when removing the state and node feedback components (averages of 3 repetitions).}
    \label{tab:ablations}
\end{table}

\section{Related Work}
\label{sec:rwork}
Our work bears semblance and owes credit to various contemporary lines of work.
From the architectural angle, we perceive our work as an application-specific offspring of weight-tied architectures, dynamic graph convolutions and structure-aware self-attention networks.
The depth recurrence of our decoder is inspired by weight-tied architectures~\cite{dehghani2018universal,bai2019deep} and their graph-oriented variants~\cite{li2016gated}, which model neural computation as the fix-point iteration of a single layer against a structured input, thus allowing for a dynamically adaptive computation ``depth'' -- albeit with a constant parameter count.
Analogously to structure-aware self-attention networks~\cite{zhu-etal-2019-modeling,cai2020graph} and graph attentive networks~\cite{velivckovic2018graph,yun2019graph,ying2021transformers,brody2021attentive}, our decoder employs standard query/key and fully-connected attention mechanisms injected with structurally biased representations, either at the edge or at the node level.
Finally, akin to dynamic graph approaches~\cite{liao2019efficient,pareja2020evolvegcn}, our decoder forms a closed loop system that autoregressively generates its own input, in the process becoming exposed to subgraph structures that drastically differ between time steps.

From the application angle, our proposal is a refinement of and a continuation to recent advances in categorial grammar supertagging.
Similar to the transition from words to subword units~\cite{sennrich-etal-2016-neural}, constructive supertaggers seek to bolster generalization by disassembling syntactic categories into smaller indivisible units, thereby incorporating structure at a finer granularity scale.
The original approach of~\citet{kogkalidis-etal-2019-constructive} employed seq2seq models to directly translate an input text to a flattened projection of a categorial sequence, demonstrating that the correct prediction of categories unseen during training is indeed feasible.
\citet{prange-etal-2021-supertagging} improved upon the process through the explicit accounting of the tree structure embedded within categorial types, while~\citet{liu2021generating} explored the orthogonal approach of employing a transition-based ``parser'' over individual categories.
Outside the constructive paradigm,~\citet{tian-etal-2020-supertagging} employed graph convolutions over sentential edges built from static, lexicon-based preferences.
Our approach is a bridge between prior works; our modeling choice of structure-aware graph convolutions boasts the merits of ex+plicit sentential and tree-structured edges, a structurally constrained, valid-by-construction output space, favorable memory and time complexities, partial auto-regressive context flows, end-to-end differentiability with no vocabulary requirements, and minimal rule-based structure manipulation.

\section{Conclusion}
We have proposed a novel supertagging methodology, where both the linear order of the output sequence and the tree-like structure of its elements is made explicit.
To represent the different information sources (sentential word order, subword contextualized vectors, tree-sequence order and intra-tree edges) 
and their disparate sizes and scales, we turned to heterogeneous graph attention networks.
To capture the auto-regressive dependencies between different trees, we formulated the task as a dynamic graph completion process, aligning each subsequent temporal step with a higher order tree node neighborhood and predicting them in parallel across the entire sequence.
We tested our methodology on four different datasets spanning three languages and as many grammar formalisms, establishing new state of the art scores in the process.
Through our ablation studies, we showed the importance of incorporating both \textit{intra}- and \textit{inter}-tree context flows, to which we attribute our system's performance.

Other than architectural adjustment and optimizations, several interesting ideas present themselves as promising research avenues.
First, it is worthwhile to consider adaptations of our framework to either allow an efficient integration of more ``exotic'' context pathways, e.g. sibling node interactions, or alter the graph's decoding order altogether. 
On a related note, for formalisms faithful to the linear logic roots of categorial grammars, it seems reasonable to anticipate that the goal graph can be compactified by collapsing primitive nodes of opposite polarity according to their interactions, unifying the tasks of supertagging and parsing with a single end-to-end framework.

Practice aside, our results pose further evidence that lexical sparsity, historically deemed the categorial grammar's curse, might well just require a change of perspective to tame and deploy as the answer to the very problem it poses.

\section*{Limitations}
Despite its objective success, our methodology is not without limitations.
Most importantly, our model trades inference speed for an incompatibility with local greedy algorithms like beam search.
Put plainly, obtaining more than the "best" category assignment per word is not straightforward, which can potentially negatively impact the downstream parser's coverage.
A possible solution would involve branching across multiple tree-slices (i.e. sequences of partial assignments) rather than single predictions, but efficiently computing scores and comparing between complex structures is uncharted territory and not trivial to implement.
Note, however, that the issue is not unique to our system but common to all decoders that perform multiple assignments concurrently.

Parallel or not, all auto-regressive decoders assume an order on their output: the standard left-to-right order (which makes sense for text) has become the de facto choice for most applications.
The order we have chosen to employ here is structurally faithful to our output, but is neither the only one, nor necessarily the most natural one.
In that sense, the entanglement between structural bias (i.e. from the graph operations and representations) and decoding priority (i.e. the order in which trees become revealed) is a practical decision rather than a deep one -- a better operationalization could for instance employ an insertion-style operation on the graph-structured output to yield an "easy-first" geometric tagger.
We await further developments and community insights on that front.

Finally, the system carries the standard risks of any NLP architecture reliant on machine learning, namely linguistic biases inherited from the unsupervised pretraining of the incorporated language models, and annotation biases derived from the supervised training over human-labeled data.

\bibliography{anthology, custom}

\appendix

\onecolumn
\section{Visualization of the decoding process}
\label{app:viz}
% Figures~\ref{fig:ar_flow}, \ref{fig:lr_flow} and \ref{fig:td_flow} present snapshots of the decoding process in the geometry-aware, left-to-right and top-down auto-regressive paradigms, focusing on the central tree $T_i$.
% The example sequence is intentionally abstract so as not to add grammar-specific overhead, but tree structure is assumed fixed and given by binary nodes $T_{i, 1}$ $T_{i, 3}$ $T_{i, 7}$ and $T_{i+1, 1}$ (rest zeroary).
% % Parallel operations are grayed out when they do not constitute immediate context for the current computation.

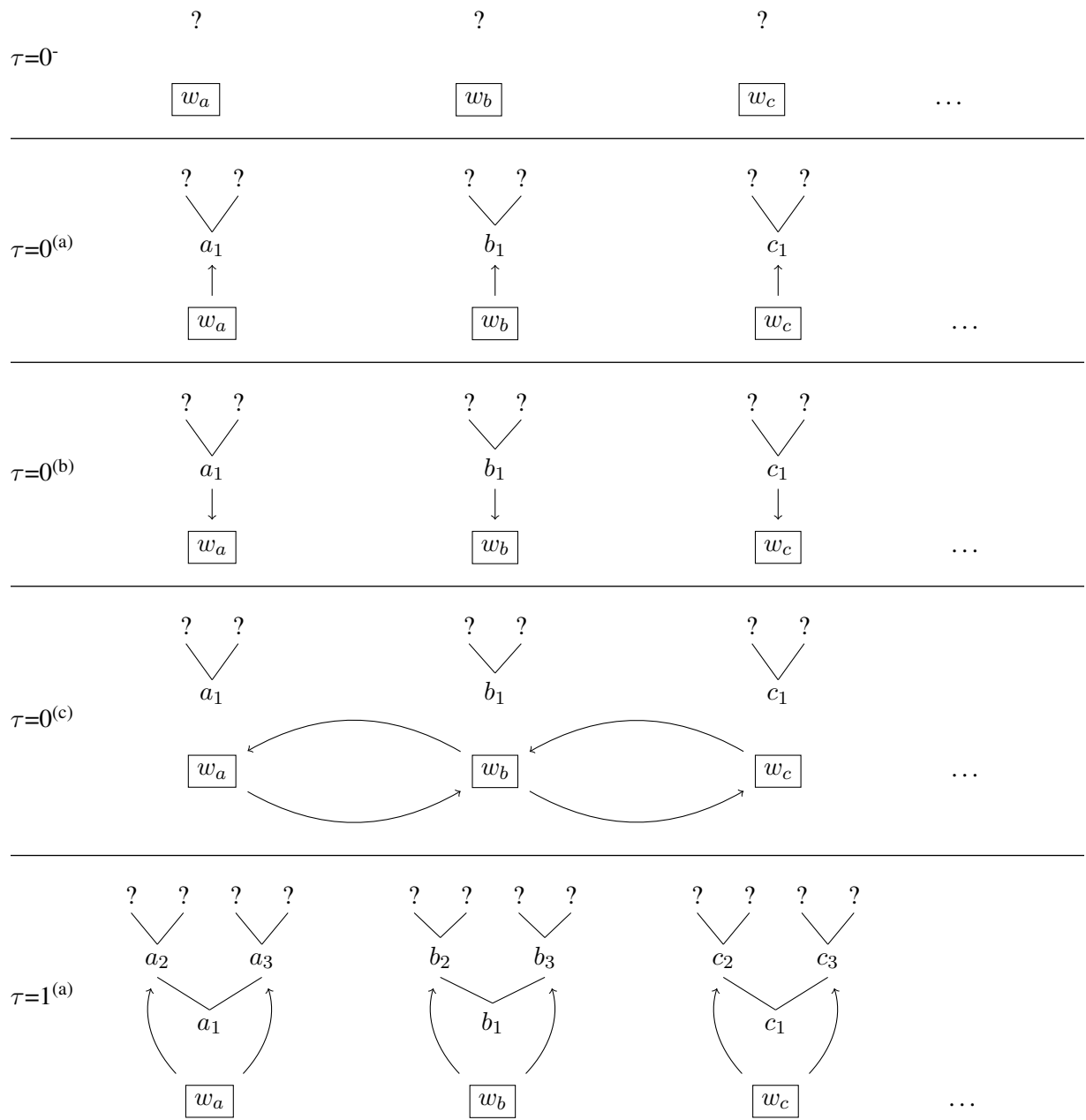
\begin{figure*}[hbtp]
	\tikzset{w/.style={draw, outer sep=5pt}}
	\tikzset{n/.style={}}
    \tikzset{grow'=up}
    \tikzset{sibling distance=10pt}
    \tikzset{level 1/.style={level distance=32pt}}
	\tikzset{level 2/.style={level distance=28pt}}
	\tikzset{level 3+/.style={level distance=26pt}}
    	$\tau$=0\textsuperscript{-}\hfill\begin{subfigure}{0.85\textwidth}
	\begin{tikzpicture}
			\begin{scope}[xshift=-120pt]
			\Tree 
				[.\node[w] {$w_a$} ;
				\edge[draw=none];
					[.{?}
					]
				]
			\end{scope}
			\Tree 
				[.\node[w] {$w_b$};
				\edge[draw=none];
					[.{?}
					]
				]
			\begin{scope}[xshift=120pt]
			\Tree 
				[.\node[w] {$w_c$};
				\edge[draw=none];
					[.{?}
					]
				]
			\end{scope}
			\begin{scope}[xshift=200pt]
				\node (dots) at (0,0) {\dots};
			\end{scope}
    \end{tikzpicture}
    \end{subfigure}\\[\smallsep]
    \rule{1\textwidth}{0.5pt}\\[\smallsep]
    	$\tau$=0\textsuperscript{(a)}\hfill\begin{subfigure}{0.85\textwidth}
	\begin{tikzpicture}
			\begin{scope}[xshift=-120pt]
			\Tree 
				[.\node[w] {$w_a$} ;
				\edge[->];
					[.{$a_1$}
						[.{?}
						]
						[.{?}
						]
					]
				]
			\end{scope}
			\Tree 
				[.\node[w] {$w_b$};
				\edge[->];
					[.{$b_1$}
						[.{?}
						]
						[.{?}
						]
					]
				]
			\begin{scope}[xshift=120pt]
			\Tree 
				[.\node[w] {$w_c$};
				\edge[->];
					[.{$c_1$}
						[.{?}
						]
						[.{?}
						]
					]
				]
			\end{scope}
			\begin{scope}[xshift=200pt]
				\node (dots) at (0,0) {\dots};
			\end{scope}
    \end{tikzpicture}
    \end{subfigure}\\[\smallsep]
    \rule{1\textwidth}{0.5pt}\\[\smallsep]
    	$\tau$=0\textsuperscript{(b)}\hfill\begin{subfigure}{0.85\textwidth}
	\begin{tikzpicture}
			\begin{scope}[xshift=-120pt]
			\Tree 
				[.\node[w] (wa) {$w_a$} ;
				\edge[<-];
					[.{$a_1$}
						[.{?}
						]
						[.{?}
						]
					]
				]
			\end{scope}
			\Tree 
				[.\node[w] (wb) {$w_b$};
				\edge[<-];
					[.{$b_1$}
						[.{?}
						]
						[.{?}
						]
					]
				]
			\begin{scope}[xshift=120pt]
			\Tree 
				[.\node[w] (wc) {$w_c$};
				\edge[<-];
					[.{$c_1$}
						[.{?}
						]
						[.{?}
						]
					]
				]
			\end{scope}
			\begin{scope}[xshift=200pt]
				\node (dots) at (0,0) {\dots};
			\end{scope}
    \end{tikzpicture}
    \end{subfigure}\\[\smallsep]
    \rule{1\textwidth}{0.5pt}\\[\smallsep]
    	$\tau$=0\textsuperscript{(c)}\hfill\begin{subfigure}{0.85\textwidth}
	\begin{tikzpicture}
			\begin{scope}[xshift=-120pt]
			\Tree 
				[.\node[w] (wa) {$w_a$} ;
				\edge[draw=none];
					[.{$a_1$}
						[.{?}
						]
						[.{?}
						]
					]
				]
			\end{scope}
			\Tree 
				[.\node[w] (wb) {$w_b$};
				\edge[draw=none];
					[.{$b_1$}
						[.{?}
						]
						[.{?}
						]
					]
				]
			\begin{scope}[xshift=120pt]
			\Tree 
				[.\node[w] (wc) {$w_c$};
				\edge[draw=none];
					[.{$c_1$}
						[.{?}
						]
						[.{?}
						]
					]
				]
			\end{scope}
			\begin{scope}[xshift=200pt]
				\node (dots) at (0,0) {\dots};
			\end{scope}
			\draw[->] (wa) to[bend right] (wb);
			\draw[->] (wb) to[bend right] (wc);
			\draw[<-] (wa) to[bend left] (wb);
			\draw[<-] (wb) to[bend left] (wc);
    \end{tikzpicture}
    \end{subfigure}\\[\smallsep]
    \rule{1\textwidth}{0.5pt}\\[\smallsep]
    	$\tau$=1\textsuperscript{(a)}\hspace{18pt}\begin{subfigure}{0.85\textwidth}
	\begin{tikzpicture}
			\begin{scope}[xshift=-120pt]
			\Tree 
				[.\node[w] (wa) {$w_a$} ;
				\edge[draw=none];
					[.{$a_1$}
						[.\node[n] (a2) {$a_2$};
							[.{?}
							]
							[.{?}
							]
						]
						[.\node[n] (a3) {$a_3$};
							[.{?}
							]
							[.{?}
							]
						]
					]
				]
			\end{scope}
			\Tree 
				[.\node[w] (wb) {$w_b$};
				\edge[draw=none];
					[.{$b_1$}
						[.\node[n] (b2) {$b_2$};
							[.{?}
							]
							[.{?}
							]
						]
						[.\node[n] (b3) {$b_3$};
							[.{?}
							]
							[.{?}
							]
						]
					]
				]
			\begin{scope}[xshift=120pt]
			\Tree 
				[.\node[w] (wc) {$w_c$};
				\edge[draw=none];
					[.{$c_1$}
						[.\node[n] (c2) {$c_2$};
							[.{?}
							]
							[.{?}
							]
						]
						[.\node[n] (c3) {$c_3$};
							[.{?}
							]
							[.{?}							]
						]
					]
				]
			\end{scope}
			\begin{scope}[xshift=200pt]
				\node (dots) at (0,0) {\dots};
			\end{scope}
			\draw[->, shorten >=5pt] (wa) to[bend left] (a2);
			\draw[->, shorten >=5pt] (wb) to[bend left] (b2);
			\draw[->, shorten >=5pt] (wc) to[bend left] (c2);
			\draw[->, shorten >=5pt] (wa) to[bend right] (a3);
			\draw[->, shorten >=5pt] (wb) to[bend right] (b3);
			\draw[->, shorten >=5pt] (wc) to[bend right] (c3);
    \end{tikzpicture}
    \end{subfigure}\\[\midsep]
    \caption{A frame by frame view of the first decoding step, where the abstract canvas assumes words $w_a$, $w_b$, $w_c$ \dots, rooting fully binary trees $a$, $b$, $c$ \dots, with nodes enumerated in a breadth-first fashion. For an intuition on what a concrete canvas might look like, refer to Figure~\ref{figure:canvas_ex}.}
    \label{figure:decoding_process}
\end{figure*}

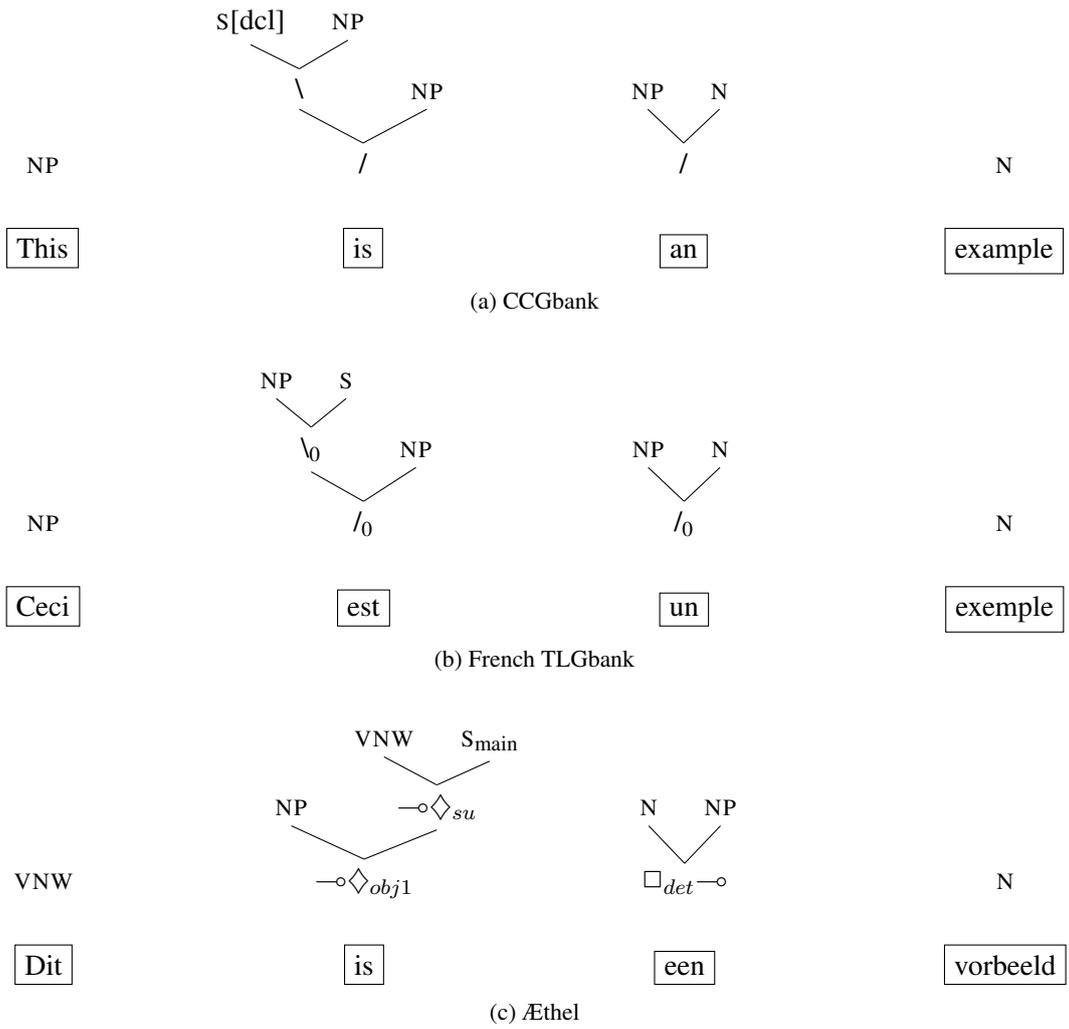
\begin{figure*}
    \tikzset{w/.style={draw, outer sep=5pt}}
	\tikzset{n/.style={}}
    \tikzset{grow'=up}
    \tikzset{sibling distance=10pt}
    \tikzset{level 1/.style={level distance=32pt}}
	\tikzset{level 2/.style={level distance=28pt}}
	\tikzset{level 3+/.style={level distance=26pt}}
	\begin{subfigure}[t]{1\textwidth}
    \centering
	\begin{tikzpicture}
			\begin{scope}[xshift=-120pt]
			\Tree 
				[.\node[w] (wa) {This} ;
				\edge[draw=none];
					\textsc{np}
				]
			\end{scope}
			\Tree 
				[.\node[w] (wb) {is};
				\edge[draw=none];
					[.{/}
						[.{\textbackslash}
						    {\textsc{s}[dcl]}
						    {\textsc{np}}
						]
					    {\textsc{np}}
					]
				]
			\begin{scope}[xshift=120pt]
			\Tree 
				[.\node[w] (wc) {an};
				\edge[draw=none];
					[.{/}
					    {\textsc{np}}
					    {\textsc{n}}
					]
				]
			\end{scope}
			\begin{scope}[xshift=240pt]
			\Tree 
				[.\node[w] (wc) {example};
				\edge[draw=none];
					\textsc{n}
				]
			\end{scope}
		\end{tikzpicture}
	\caption{CCGbank}
	\end{subfigure}\\[\midsep]
	\begin{subfigure}[t]{1\textwidth}
    \centering
	\begin{tikzpicture}
			\begin{scope}[xshift=-120pt]
			\Tree 
				[.\node[w] (wa) {Ceci} ;
				\edge[draw=none];
					\textsc{np}
				]
			\end{scope}
			\Tree 
				[.\node[w] (wb) {est};
				\edge[draw=none];
					[.{/\textsubscript{0}}
						[.{\textbackslash\textsubscript{0}}
						    {\textsc{np}}
						    {\textsc{s}}
						]
					    {\textsc{np}}
					]
				]
			\begin{scope}[xshift=120pt]
			\Tree 
				[.\node[w] (wc) {un};
				\edge[draw=none];
					[.{/\textsubscript{0}}
					    {\textsc{np}}
					    {\textsc{n}}
					]
				]
			\end{scope}
			\begin{scope}[xshift=240pt]
			\Tree 
				[.\node[w] (wc) {exemple};
				\edge[draw=none];
					\textsc{n}
				]
			\end{scope}
		\end{tikzpicture}
	\caption{French TLGbank}
	\end{subfigure}\\[\midsep]
	\begin{subfigure}[t]{1\textwidth}
    \centering
	\begin{tikzpicture}
			\begin{scope}[xshift=-120pt]
			\Tree 
				[.\node[w] (wa) {Dit} ;
				\edge[draw=none];
					\textsc{vnw}
				]
			\end{scope}
			\Tree 
				[.\node[w] (wb) {is};
				\edge[draw=none];
					[.{${\scalebox{1}[1]{\ensuremath{\multimap}}}\diamondsuit_{obj1}$}
					    {\textsc{np}}
						[.{${\scalebox{1}[1]{\ensuremath{\multimap}}}\diamondsuit_{su}$}
						    {\textsc{vnw}}
						    {\textsc{s}\textsubscript{main}}
						]
					]
				]
			\begin{scope}[xshift=120pt]
			\Tree 
				[.\node[w] (wc) {een};
				\edge[draw=none];
					[.{$\Box_{det}{\scalebox{1}[1]{\ensuremath{\multimap}}}$}
					    {\textsc{n}}
					    {\textsc{np}}
					]
				]
			\end{scope}
			\begin{scope}[xshift=240pt]
			\Tree 
				[.\node[w] (wc) {vorbeeld};
				\edge[draw=none];
					\textsc{n}
				]
			\end{scope}
		\end{tikzpicture}
	\caption{\AE thel}
	\end{subfigure}
	\caption{Artificial but concrete canvas examples for the three grammars experimented on.}
	\label{figure:canvas_ex}
\end{figure*}

% \clearpage

\end{document}